\documentclass[amsmath,amssymb]{revtex4}\usepackage{graphicx}
\usepackage{color}
\usepackage{dcolumn}
\usepackage{bm}
\usepackage{amsmath,amsthm}
\usepackage{rotating}%contains sidewaystable

\DeclareMathOperator{\NGD}{\mathbf{NGD}}
\begin{document}
	\preprint{APS/123-QED}
	\title{\textsc{Google Distance Between Words }}
	\author{Alberto J. Evangelista}
	\email{Alberto.J.Evangelista.01@alum.dartmouth.org}
	\affiliation{Department of Physics\\ University of Connecticut\\2152 Hillside Road, Unit 3046\\Storrs, CT 06269}
	\author{Bj\o rn Kjos-Hanssen}
	\email{bjoernkh@hawaii.edu}
	\thanks{this work was partially supported by a grant from the Simons Foundation (\#315188 to Bj\o rn Kjos-Hanssen).}
	\affiliation{Department of Mathematics\\ University of Hawai\textquoteleft i at M\=anoa\\ 2565 McCarthy Mall \\ Honolulu, HI 96822}
	\date{\today}
	\begin{abstract}
		Rudi Cilibrasi and Paul Vit\'anyi have demonstrated that it is possible to extract the meaning of words from the world-wide web.
		To achieve this, they rely on the number of webpages that are found through a Google search containing a given word and they associate the page count to the probability that the word appears on a webpage.
		Thus, conditional probabilities allow them to correlate one word with another word's meaning. Furthermore, they have developed a similarity {distance function} that gauges how closely related a pair of words is.
		We present a specific counterexample to the triangle inequality for this similarity distance function.
	\end{abstract}
	\pacs{Valid PACS appear here}
	\maketitle
	\section{Introduction}
		When the Google search engine is used to search for word $x$, Google displays the number of hits that word $x$ has.
		The ratio of this number of hits to the total number of webpages indexed by Google represents the probability that word $x$ appears on a webpage.
		Cilibrasi and Vit\'anyi \cite{CV} use this probability to extract the meaning of words from the world-wide-web.
		If word $y$ has a higher conditional probability to appear on a webpage, given that word $x$ also appears on the webpage, than it does by itself, then it can be concluded that words $x$ and $y$ are related.
		Moreover, higher conditional probabilities imply a closer relationship between the two words.
		Thus, word $x$ provides some meaning to word $y$ and vice versa.

		 Cilibrasi and Vit\'anyi's \emph{normalized Google distance} (NGD) function measures how close word $x$ is to word $y$ on a zero to infinity scale.
		A distance of zero indicates that the two words are practically the same.
		Two independent words have a distance of one.
		A distance of infinity occurs for two words that never appear together.

		Although Cilibrasi and Vit\'anyi's NGD function was sensibly derived from its basic axioms (which allow it to theoretically yield the values mentioned above), it does not account for the presence of multi-thematic webpages.
		In other words, the NGD function does not account for webpages, such as dictionary sites and other long pages, which encompass many unrelated subjects.
		For this reason, it is necessary to renormalize the NGD formula to achieve the desired values. 

	\section{NGD's expectation value}
	
		On the average, two random words should be independent of one another.
		Hence, two random words should have an NGD of one.
		To test this assumption, we randomly selected two sets of five words from the dictionary.
		In addition, we randomly chose one set of ten words from two different news articles (five words were taken from each article).
		We then proceeded to evaluate the NGD among the different word pairs in each set using Cilibrasi and Vit\'anyi's formula
		\begin{equation}
			\NGD(x,y)
			= \frac{\max \left\{\log f(x),\log f(y)\right\} - \log f(x,y)}
			{{\log M - \min \left\{\log f(x),\log f(y)\right\}}}, 
		\end{equation}
		where $f(x)$ and $f(y)$ are the number of hits of words $x$ and $y$, respectively, and $M$ is the total number of webpages that Google indexes.

		Our first set, which consisted of the words \texttt{micrometeorite}, \texttt{transient}, \texttt{denature}, \texttt{pentameter}, and \texttt{reside}, yielded an expectation value of 0.64 for the NGD with a standard deviation of 0.14.
		Our second set, which consisted of the words \texttt{detrition}, \texttt{unity}, \texttt{interstice}, \texttt{abrupt}, and \texttt{reside}, had an expectation value of 0.75 for the NGD with a standard deviation of 0.12.
		Our last set, which consisted of the words \texttt{agency}, \texttt{diabetic}, \texttt{enforcement}, \texttt{federal}, \texttt{hormone}, \texttt{illegal}, \texttt{intelligence}, \texttt{measure}, \texttt{spread}, and \texttt{war},
		yielded an expectation value of 0.77 with a standard deviation of 0.15.
		Averaging the above values (weighing them correctly according to the number of words in each set), we obtain an expected value of 0.7325 for the NGD.
		A similar evaluation provides us with a standard deviation of 0.14.

		A different type of analysis brought us to a similar expectation value.
		We call this evaluation the \emph{triangle difference}, (\textbf{TD}).
		The reason for this name is that we evaluated the following difference:
		\begin{equation}
			{\mathbf{TD}} \equiv \NGD(x,y) + \NGD(y,z) - \NGD(x,z).
		\end{equation}
		The reason to evaluate such difference stemmed from the possibility that the sum of two distances, between words x and y, and y and z, might be smaller than the distance between x and z.
		If such were the case, then it would be sensible to redefine the distance between two words such that it minimizes all possible NGD sums:
		\begin{equation}
			\NGD^{*}(x,z) \equiv \min \left\{ \NGD(x,y) + \NGD(y,z),\NGD(x,z)\right\},
		\end{equation}
		over all words $y$. Of course, it is not practical to consider all possible $y$.
		The triangle difference, however, is only violated by extremely rare exceptions, so it is not necessary to perform such minimization.
		Nevertheless, we proceeded to evaluate the expected triangle difference for each of our sets.
		They were 0.69, 0.79, and 0.60, for the first, second, and third sets, respectively.
		Combining them, they yield an expected triangle difference of 0.67.
		This, of course, is close to the expected value of NGD, as the triangle difference for random words is
		\begin{equation}
			E\left[{\mathbf{TD}}\right] \equiv E\left[\NGD(x,y)\right] + E\left[\NGD(y,z)\right]
		 	- E\left[\NGD(x,z)\right] = 2E\left[\NGD\right] - E\left[\NGD\right] = E\left[\NGD\right].
		\end{equation}
		Therefore, the expectation value of the triangle difference should be equal to the expectation value of the NGD.
		A rough average of our expectation values obtained through each method is 0.7.

	\section{Notes Regarding NGD}
		In an arduous effort to find a set of words that would violate the triangle difference, we obtained a set that illustrates a few interesting properties of NGD.
		The set consists of the words \texttt{Rolling Stones}, \texttt{Beatles}, and \texttt{salmonflies} and it is, among the many word sets that we attempted, the only one that violates the triangle difference.
		Our first evaluation of the pertinent distances is given as $\NGD_1$ in Figure \ref{1}.

		\begin{figure}
			\begin{tabular}{c c c}
				Word Pair & $\NGD_1$ & $\NGD_2$\\
				\hline
				\texttt{Rolling Stones}, \texttt{Beatles}		& 0.23 	& 0.27\\
				\texttt{Beatles}, \texttt{salmonflies}			& 0.81  & 0.82\\
				\texttt{Rolling Stones}, \texttt{salmonflies} 	& 1.06  & 1.14\\
			\end{tabular}
			\caption{Two evaluations of NGD values among the words \texttt{Rolling Stones}, \texttt{Beatles}, and \texttt{salmonflies} in the year 2006.}\label{1}
		\end{figure}

		\noindent As can be observed, the NGD between \texttt{Rolling Stones} and \texttt{salmonflies} (1.06) is slightly higher than the addition of the Google distances between \texttt{Rolling Stones} and \texttt{Beatles},
		and between \texttt{Beatles} and \texttt{salmonflies} (1.04).
		Therefore, our first observed property of NGD is that, even in the rare cases in which the triangle difference is violated, it is not by much.
		Furthermore, it is important to indicate that our example worked because of the high propensity that people have to misspell the word beetle as beatle, thus decreasing the distance between \texttt{Beatles} and \texttt{salmonflies}.

		The second property that we observed, which deserves much attention, is the NGD dependence on the Google server to which a user connects.
		A second evaluation of the distances in question yielded the result shown as $\NGD_2$ in Figure \ref{1}.

		\noindent This second set of distances was obtained by connecting to Google through a different internet service provider and it shows that Google distances are not stable values.
		In fact, from our example, they can vary by as much as 17\% (for the \texttt{Rolling Stones-Beatles} word pair).

		The last property that this set of words depicts is the plausibility of Google distances that are higher than unity.
		With an NGD of 1.14, the word pair \texttt{Rolling Stones-salmonflies} has the highest distance that we have encountered.
		Indeed, Google distances greater than one are very rare, and before this example, the only one we had encountered was between the words \texttt{transient} and \texttt{pentameter} (1.02).
		Moreover, these cases are so rare that even Cilibrasi and Vit\'anyi's conjecture about the words \texttt{by} and \texttt{with} having an NGD higher than one is false.
		The actual distance, per the Google server currently in use, is 0.19.

	\section{Conclusions}
		Although Google can be used to extract the meaning of words, it is important to modify Equation 1 in order to obtain the desired distance values between words.
		The expectation value of NGD, which is the distance between two random and therefore independent words, is 0.7.
		To achieve the desired value of unity between independent words, it is only necessary to recalibrate the NGD formula by dividing by 0.7:
		\begin{equation}
			\NGD^{*}(x,y) =
			\frac{\NGD(x,y)}{{0.7}}.
		\end{equation}

		It is also important to remember that NGD values are not exact.
		They depend on the number of hits that each word has, which makes them unstable.
		Factors such as the Google server to which one connects and the number of websites connected to the world-wide-web can cause discrepancies as high as 17\%, which our \texttt{Rolling Stones-Beatles} example showed.

	\appendix*\section{From 2006 to 2015}	
		The research reported above was conducted in 2006 and an arXiv version has been cited by several researchers.
		For consistency with that version we separately provide an up-to-date view for 2015 in this appendix.
		It would be interesting to reconstruct the Google search engine as it existed in 2006, but we are not in a position to do so.
		\subsection{Accuracy of reported number of results}
			Between the date of this research in 2006 and the current date of 2015,
			Google's search engine has changed. The reported numbers of results are now clearly not really reliable. Adding an extra keyword can increase, rather than decrease, the reported number of results. See for instance the example of adding ``-used'' to the search ``car''\footnote{{http://searchengineland.com/why-google-cant-count-results-properly-53559}}.

			Let us write $g(x)=n$ if the Google search engine reports that there are
			\begin{quote}
				``About $n$ results''
			\end{quote}
			for the query $x$.
			As an egregious example, we found that
			\[
				g(\texttt{"clinton"})= 239 \text{ million},\quad g(\texttt{"bush"})= 373 \text{ million}, 
			\]
			\[
				g(\texttt{"clinton" OR "bush"}) = 607 \text{ million},\quad\text{and }
				g(\texttt{"clinton" "bush"})= 151 \text{ million}.
			\]
			which clearly violates basic counting.

			Cohen and Vit\'anyi \cite[page 16]{DBLP:journals/corr/CohenV13} nevertheless state that
			\begin{quote}
				``[...] we have found that these approximate
				measures are sufficient to generate useful answers [...]''
			\end{quote}
		\subsection{Failure of the triangle inequality}
			Cohen and Vit\'anyi \cite[Remark IV.4]{DBLP:journals/corr/CohenV13} state
			\begin{quote}
				``In practice for the combination of the World Wide Web and Google the NGD
				may satisfy the triangle inequality. We did not find a counterexample.''
			\end{quote}
			We now find that our 2006 example no longer violates the triangle inequality. This seems to be due to the existence of an arXiv version of our paper on the web which mentions this specific triple.
			Thus we set out to find a new example.

			Our methodology was to enter the words into the Google search interface, during the period December 29, 2014 through January 2, 2015, in the form
			\begin{quote}
				\texttt{"cojones"}
			\end{quote}
			for a single word, and
			\begin{quote}
				\texttt{"cojones"} \texttt{"separability"}
			\end{quote}
			for a pair of words.
			The quotes ensure that only exact matches will be considered.
			We used
			\[
				M = g(\texttt{the}) = 25,270,000,000,
			\]
			the number of hits for the very common word \texttt{the}, following \cite[Section 1.A]{DBLP:journals/corr/CohenV13}.

			\begin{table}
				\begin{tabular}{c|r}
				\hline
				$g(\texttt{"glen"})$ &268,000,000\\
				$g(\texttt{"alien"})$ &278,000,000\\
				$g(\texttt{"torrid"})$ & 10,400,000\\
				\hline
				$g(\texttt{"glen" "alien"})$&843,000\\
				$g(\texttt{"glen" "torrid"})$&6,950,000\\
				$g(\texttt{"alien" "torrid"})$&11,400,000\\
				\hline
				$\NGD(\texttt{glen},\texttt{alien})$	&	1.28\\
				$\NGD(\texttt{glen},\texttt{torrid})$	&	0.47\\
				$\NGD(\texttt{alien},\texttt{torrid})$	&	0.41\\
				\hline
				\end{tabular}
				\caption{A counterexample for the triangle inequality on January 2, 2015.}\label{tehuset}
			\end{table}

			To find an example, we drew 50 random triples from the \emph{Corpus of Contemporary American English} (COCA) \cite{COCA} using the ``random'' button of its online interface. Let $X$ denote the number of counterexamples to the triangle inequality in such a sample. Thus, $X$ is a binomial random variable with a parameter $n=50$ and an unknown parameter $p$. We observed $X=0$.

			Rather than increasing the sample size, we next employed a simple kind of \emph{simulated annealing} algorithm. That is, we selected the pair of words in the sample having the highest distance: $\NGD(\texttt{glen}, \texttt{alien})=1.28$.
			Then we replaced the third word in the triple by another random word, expecting to have to repeat the replacement several times.
			However, it immediately gave an example; see Table \ref{tehuset}.

			Having found a convincing example of the failure of the triangle inequality, we may wonder how common this phenomenon is. Our results suggest that roughly 1 in
			\[
				200=(1+3)\times 50
			\]
			triples from the COCA fail to satisfy the triangle inequality. (Each original triple is resampled in 3 ways depending on which pair is chosen.)

			On the other hand, the fact that we obtained $X=0$ means that we can assert with 95\% confidence that $p\le 1/16$.
			Indeed, if $X$ has parameter $p\ge 1/16$ then
			\[
				\mathbb P(X=0) \le \mathbb P(X = 0\mid p=1/16) = (1-1/16)^{50}=3.97\%<5\%.
			\]

			The results for the first 10 triples are shown in Table \ref{referee}. Already in this small set of 10 triples we notice problems: \texttt{burnished} returns more results when combined with \texttt{growth} than by itself, and \texttt{capstan} has the same problem when combined with \texttt{fashion}.

			The mean and sample standard deviation of the $\NGD$ for the three subsamples of first, second, and third element of each triple are shown in Table \ref{SFO}.

			\begin{table}
				\begin{tabular}{c|c|c}
					Sample & $\texttt{AVERAGE}(\NGD)$ & $\texttt{STDEV}(\NGD)$\\
					\hline
					 1 &0.59&	0.20\\
					 2 &0.61&	0.16\\
					 3 &0.57&	0.14\\
				\end{tabular}
				\caption{Mean and standard deviation.}\label{SFO}
			\end{table}

			\begin{sidewaystable}
				\begin{tabular}{|r|r|r|r|r|r|r|r|r|r|r|}
				\hline
				$A$	&	cojones	&	airtime	&	growth	&	harmonized	&	silky-smooth	&	poor-performing	&	business-oriented	&	semi-literate	&	indignity	&	documentary-style\\
				$B$	&	separability	&	waxing	&	klutzy	&	materialize	&	flub	&	big-company	&	wheelbarrow	&	parenthetical	&	carter & capstan\\
				$C$ & retirement & locomotion & burnished & swarming & unfunny & artilleryman & senselessness & mankind & ceremonial & fashion\\
				\hline
				\hline
				$g(\texttt{"}A\texttt{"})$ &  9,850,000&	 9,390,000&	805,000,000&	8,950,000&	1,640,000&	  392,000&	  881,000&	   424,000&	  1,530,000&	      548,000\\
				$g(\texttt{"}B\texttt{"})$ &    641,000&	46,400,000&	    409,000&	7,940,000&	  538,000&	2,830,000&	7,840,000&	   623,000&	372,000,000&	      744,000\\
				$g(\texttt{"}C\texttt{"})$ &226,000,000&	10,700,000&	 10,500,000&	6,170,000&	  851,000&	  353,000&	  389,000&	81,800,000&	 27,500,000&	1,690,000,000\\
				\hline
				$g(\texttt{"}A\texttt{"} \texttt{"}B\texttt{"})$&        566&	   428,000&	    233,000&	  529,000&	    7,320&	    4,030&	   14,000&	     3,150&	  2,120,000&	        1,290\\
				$g(\texttt{"}A\texttt{"} \texttt{"}C\texttt{"})$&    968,000&	   111,000&	 10,700,000&	  378,000&	   12,700&	       45&	      963&	   109,000&	    741,000&	      206,000\\
				$g(\texttt{"}B\texttt{"} \texttt{"}C\texttt{"})$&    380,000&	   718,000&	      7,130&	  425,000&	   25,800&	      557&	    7,130&	   527,000&	 13,700,000&	    8,340,000\\
				\hline
				\hline
				$\NGD(A,B)$&0.92&	0.59&	0.74&	0.35&	0.50&	0.59&	0.62&	0.48&	0.53&	0.59\\
				$\NGD(A,C)$&0.69&	0.58&	0.55&	0.38&	0.47&	0.81&	0.62&	0.60&	0.37&	0.84\\
				$\NGD(B,C)$&0.60&	0.54&	0.66&	0.35&	0.32&	0.76&	0.63&	0.48&	0.48&	0.51\\
				\hline
				\end{tabular}
				\caption{Testing the triangle inequality.}\label{referee}
			\end{sidewaystable}

			Since the $\NGD$ does not satisfy the triangle inequality, we could consider the following alternative to the $\NGD$.
			The \emph{normalized symmetric-set-difference metric} is defined by
			\[
				d(A,B) = \mathbb P(\overline A\cup\overline B\mid A\cup B),
			\]
			where $\overline A$ denotes the complement of $A$.
			It has the advantage of satisfying the triangle inequality and not needing an overall universe size $M$.
			It is bounded above by 1, which is achieved for disjoint sets $A$ and $B$.
			The fact that $d$ satisfies the triangle inequality is not immediately obvious;
			it is proved by P.\ Yianilos \cite{Yianilos91}.

	\begin{acknowledgments}
		We thank the Office of Undergraduate Research, University of Connecticut, for selecting our work for presentation \cite{EK}.
		The second author thanks the organizers M.~Hutter, W.~Merkle and P.~Vit\'anyi for the invitation to participate in the IBFI workshop
		\emph{Kolmogorov Complexity and Its Applications}, Schlo{\ss} Dagstuhl, Germany, January--February 2006.
	\end{acknowledgments}
	\bibliography{Evangelista_Kjos-Hanssen}
\end{document}